\documentclass[letterpaper]{article} % DO NOT CHANGE THIS
\usepackage{aaai25}  % DO NOT CHANGE THIS
\usepackage{times}  % DO NOT CHANGE THIS
\usepackage{helvet}  % DO NOT CHANGE THIS
\usepackage{courier}  % DO NOT CHANGE THIS
\usepackage[hyphens]{url}  % DO NOT CHANGE THIS
\usepackage{graphicx} % DO NOT CHANGE THIS
\usepackage{natbib}  % DO NOT CHANGE THIS AND DO NOT ADD ANY OPTIONS TO IT
\usepackage{caption} % DO NOT CHANGE THIS AND DO NOT ADD ANY OPTIONS TO IT
\usepackage{algorithm}
\usepackage{algorithmic}
\usepackage{multirow}
\usepackage{scrextend}

\usepackage{newfloat}
\usepackage{listings}
\DeclareCaptionStyle{ruled}{labelfont=normalfont,labelsep=colon,strut=off} % DO NOT CHANGE THIS
\floatstyle{ruled}
\newfloat{listing}{tb}{lst}{}
\floatname{listing}{Listing}
\usepackage{todonotes}
\nocopyright % Your paper will not be published if you use this command
\usepackage{amsfonts}
\usepackage{amsmath}
\usepackage{adjustbox}
\usepackage{booktabs}

\title{Small Graph Is All You Need: DeepStateGNN for Scalable Traffic Forecasting}
\author {
    Yannick Wölker\textsuperscript{\rm 1}\equalcontrib, 
    Arash Hajisafi\textsuperscript{\rm 2}\equalcontrib, 
    Cyrus Shahabi\textsuperscript{\rm 2}, 
    Matthias Renz\textsuperscript{\rm 1}
}
\affiliations {
    \textsuperscript{\rm 1}Department of Computer Science, Kiel University, Kiel, Germany\\
    \{ywoe, mr\}@informatik.uni-kiel.de\\
    \textsuperscript{\rm 2}Department of Computer Science, University of Southern California, Los Angeles, USA\\
    \{hajisafi, shahabi\}@usc.edu
}

\begin{document}

\maketitle

\begin{abstract}
We propose a novel Graph Neural Network (GNN) model, named DeepStateGNN, for analyzing traffic data, demonstrating its efficacy in two critical tasks: forecasting and reconstruction. 
Unlike typical GNN methods that treat each traffic sensor as an individual graph node, DeepStateGNN clusters sensors into higher-level graph nodes, dubbed Deep State Nodes, based on various similarity criteria, resulting in a fixed number of nodes in a Deep State graph. 
The term “Deep State” nodes is a play on words, referencing hidden networks of power that, like these nodes, secretly govern traffic independently of visible sensors. These Deep State Nodes are defined by several similarity factors, including spatial proximity (e.g., sensors located nearby in the road network), functional similarity (e.g., sensors on similar types of freeways), and behavioral similarity under specific conditions (e.g., traffic behavior during rain). 
This clustering approach allows for dynamic and adaptive node grouping, as sensors can belong to multiple clusters and clusters may evolve over time. 
Our experimental results show that DeepStateGNN offers superior scalability and faster training, while also delivering more accurate results than competitors. It effectively handles large-scale sensor networks, outperforming other methods in both traffic forecasting and reconstruction accuracy.
% Our experimental results show that DeepStateGNN significantly improves scalability and performance, efficiently handling large-scale sensor networks with improved computation time and accuracy for both traffic forecasting and reconstruction tasks.

\end{abstract}

% Uncomment the following to link to your code, datasets, an extended version or similar.
%
% \begin{links}
%     \link{Code}{https://aaai.org/example/code}
%     \link{Datasets}{https://aaai.org/example/datasets}
%     \link{Extended version}{https://aaai.org/example/extended-version}
% \end{links}

\section{Introduction}
% ======== Motivation ==========
Traffic flow forecasting is a critical function in spatiotemporal forecasting research, providing insights to guide infrastructure development, enhance safety, improve traffic management, and integrate multimodal transportation systems. 
% While many transportation agencies traditionally rely on loop-detector sensors to gather aggregated measurements at specific time intervals~\cite{anastasiou19}, this is just one of several data sources available. The collected data, regardless of the sensor type, results in time series that present a multivariate time-series forecasting challenge.
Presently, traffic data is collected either through loop-detector sensors installed on roads or through trajectory data from vehicles, which is often transformed into traffic flow at discrete points along the road, effectively creating virtual sensors\footnote{Hereafter, the term “sensor” will refer to both real and virtual sensors.}~\cite{li2020trajectory}. Regardless of the acquisition method, this aggregated traffic data forms time series, making traffic forecasting a multivariate time-series forecasting task.

This spatiotemporal forecasting task is challenging due to complex spatiotemporal dependencies across sensors and issues with long-term dependencies in the sensors themselves. 
Previous studies have shown that graphs are effective in capturing these complex relationships~\cite{li2018}, leading to the use of graph-based approaches for traffic forecasting. 
These approaches model inter-sensor dependencies as a sensor graph, capturing relationships through message-passing techniques. 
However, there are shortcomings, three of which we highlight here. 

First, as the number of sensors increases to cover larger areas, the sensor graph grows so large that memory and computation requirements become infeasible. Consequently, these approaches are typically applied to datasets containing only highway sensors, excluding smaller arterial roads, which are crucial for analyzing traffic flow. %Analyzing PeMS data, we found that there are up to twenty times more arterial road sensors than highway sensors in Los Angeles.%, making it infeasible for previous approaches to handle such datasets.

Second, these approaches rely on a static set of sensors and require continuous data from all sensors, making them inflexible when sensors are added, removed, or experience failures. In production environments, traffic sensors can experience outages, leading to missing data. Prominent benchmark datasets like METR-LA and PEMS-BAY \cite{li2018} preprocess their data to include only sensors that are consistently available. Some studies explore imputation as a method to fill these gaps, where missing sensor data is treated similarly to traffic prediction.
%: zeros are inserted in the input sequence as placeholders, and the output is the imputed traffic values. 
However, these methods are brittle with respect to the sensor network and struggle with long-term changes in sensor layouts. When cities update their sensor layouts or add new streets with sensors, even imputation-based techniques require retraining and expanding the sensor graph.

Third, most previous GNN-based approaches focus on a single type of road, typically highways, when modeling traffic. However, different road types exhibit different traffic patterns. For example, highways show strong traffic signal propagation along their length~\cite{Pan22}, resulting in strong auto-correlations. In contrast, urban arterial roads exhibit more diverse traffic patterns. Shao et al.~\cite{shao22} highlight the importance of modeling different road classes for effective traffic forecasting, but prior work has often neglected this due to the reliance on benchmark datasets that contain only highway data.

% (\textbf{TODO: Some example from previous GNNs like DCRNN})
% Third, due to the nature of highways traffic signals strongly propagate along the roads \cite{Pan22}. 
% In a city, arterial roads are a major part of connections that will have more diverse signals similar to highway ramps traffic. This was approached by Shao \cite{shao22} to decode the explicit ramp signal from the highway sensors. 
% Third, highways exhibit strong traffic signal propagation along their length~\cite{Pan22} resulting in strong auto-correlations. In contrast, in urban areas, arterial roads play a major role in connectivity and exhibit more diverse traffic patterns.
% First work by Shao \cite{shao22} addressed this on pure freeway dataset by decoding explicit ramp signals from highway sensors, highlighting the importance to model different road classes for successful traffic forecasting.

To tackle these shortcomings, our core idea is to group similar sensors for a more effective and reliable representation instead of using a traditional sensor graph with one node for each sensor. However, such an aggregated view often leads to information loss, and how to group sensors effectively remains a challenge. These groupings should be dynamic and non-exclusive, as sensor similarities change depending on time and context. For example, sensors in different parts of a city may show similar patterns during rainy rush hours but differ on sunny holiday afternoons. Learning to group sensors and represent each sensor through a combination of observed traffic patterns from such groups can provide a fixed-size representation that scales easily and handles changes in sensor availability.

Consequently, we introduce the DeepState Graph Neural Network (DeepStateGNN) framework, which utilizes a fixed-size graph, called the DeepState graph. In this graph, each node, dubbed a DeepState Node (DSN), represents groups of sensors rather than individual sensors. These groupings are dynamically formed based on external factors such as spatial location, time, environmental conditions, or the similarity of traffic patterns captured by sensors. Each DSN aggregates the information from its corresponding sensor group into a latent state. The term ``Deep State'' is a playful nod to the concept of hidden networks of influence, mirroring how these nodes covertly form powerful clusters that govern the state of traffic independently of the individual sensor nodes, much like a ``deep state'' in political discourse operates behind the scenes.

The relationships between DSNs, both long-term and short-term, are represented as edges in the DeepState graph. 
%This is captured by inferring the similarities between the immediate latent states and global-level DSN embeddings shared across all timestamps, respectively. 
A message-passing operation on the DeepState graph allows DSNs to exchange information about the traffic state with other related DSNs. 
%The resulting DSN states effectively provide a decomposition of the traffic observations. 
Intuitively, this decomposition allows each DSN to specialize in representing specific traffic patterns, like traffic during rainy days or in particular neighborhoods. A traffic observation can be seen as a combination of these specialized nodes. Thus, this decomposition enables the inference of unknown traffic states at any location in the road network by querying a combination of DSN states that are semantically close to the query for the given time window.

Overall, our main contributions are as follows:
\begin{itemize}
    \item We introduce DeepStateGNN, a novel framework that learns a fixed-size graph representation that captures the latent traffic state for groups of similar sensors during a time window. This graph can be queried to forecast or reconstruct traffic data for locations with missing or not yet observed sensors.
    \item Our approach, with a fixed number of high-level nodes in the graph representation, is flexible and can handle datasets where sensors are added, removed, or relocated. This adaptability allows for greater data utilization and reduces the need for frequent retraining in real-world applications.
    % \item The proposed architecture outperforms all previous state-of-the-art baselines in both traffic forecasting and reconstruction accuracy, and demonstrates better computational scalability.
    \item The proposed architecture outperforms all previous state-of-the-art baselines in both traffic forecasting and reconstruction accuracy, while also demonstrating superior computational scalability, achieving both without compromising one for the other.
    \item We propose and publish the METR-LA+ traffic dataset, sourced from the same data as METR-LA and PeMS-Bay~\cite{li2018}. This dataset provides an in-the-wild representation of real-world traffic conditions compared to previous curated datasets. It includes two months of traffic data from both freeways and nearby arterial roads, incorporates missing sensors, and is enriched with additional information such as weather, air quality, and road semantics, for the same study area, tagged with location and time.
\end{itemize}

We validate our contributions through three stages of experiments. First, we show that DeepStateGNN outperforms state-of-the-art baselines on both freeway sensors and the broader network that includes arterial roads, with improvements ranging up to 40\% across all metrics. We evaluate the generalization of our approach on two tasks: traffic prediction and reconstruction. Second, we demonstrate the scalability of DeepStateGNN, showing that our fixed-size graph representation results in better scaling of training times compared to those of the baselines. Finally, an ablation study validates the effectiveness of our DeepState nodes and the design choices regarding the graph-based representation.

\section{Related Work}\label{sec:related-work}
Traffic forecasting models often rely on data from static sensors, such as loop detectors or trajectory-based flows, which are geo-referenced and map-matched on the road network. This makes them well-suited for graph-based modeling. Early methods, such as STGCN \cite{yu18} and DCRNN \cite{li2018}, used road network distances to construct static adjacency matrices. 
These static adjacency matrices capture street directionality but require many graph convolutional network (GCN) \cite{kipf2017semisupervised} steps to propagate across large network areas.
To address the limitations of static adjacency matrices, \cite{Wu19} introduced Graph WaveNet, which employs a self-adaptive adjacency matrix optimized through stochastic gradient descent. 
This approach does not rely on prior road network information; instead, it uses static node embeddings, with the transition matrix formed by applying softmax to the product of these embeddings.
A3T-GCN \cite{Bai21} leverages an attention mechanism to generate attention scores across different timestamps, while maintaining a static, binary adjacency matrix connecting direct neighbors.
With the rise of graph attention mechanisms like GAT \cite{Velickovic18}, attention-based techniques were incorporated into spatiotemporal traffic forecasting. 
%GMAN \cite{Zheng20} introduced Spatial Attention Blocks, allowing attention to be computed between any two sensors. 
%This approach captures rapidly changing spatial correlations, utilizing spatial embeddings based on node2vec and temporal embeddings derived from temporal context.
D2STGNN \cite{shao22} utilizes this concept by creating dynamic transition matrices with a self-attention mechanism. It leverages historical traffic data, temporal context, and static node embeddings, similar to Graph WaveNet, to produce context-dependent transition matrices. Although this method is more dynamic, it is tied to a sensor graph.
%, and its adjacency matrix is defined by input signals, which is a challenge handling missing data. 
Moreover, all of these methods treated sensors as graph nodes, limiting their scalability and, as a result, focusing solely on a small, curated set of freeway sensors provided by previous studies.

Beyond traffic forecasting, only two recent studies have employed the concept of high-level graph nodes. The first, BysGNN \cite{Hajisafi23}, predicts visitor numbers for Points of Interest (POIs) by modeling them as a graph with meta nodes representing clusters of similar POIs. The dynamic adjacency matrix is a function of spatial distance, temporal, and semantic embeddings. However, these meta nodes are manually crafted, static, and coexist with POIs in the graph, rather than fully representing them.
The second work, SUSTeR \cite{woelker23}, uses a graph with only abstract nodes. The adjacency matrix is derived from node embeddings based on assigned sensors. However, SUSTeR employs a static, learned assignment from sensors to abstract nodes based solely on spatial features, without dynamic grouping.
Therefore, the SUSTeR focused only on reconstruction and not forecasting. In Section~\ref{sec:experiment}, we compare DeepStateGNN with both BysGNN and SUSTeR.

\section{Preliminaries}\label{sec:preliminaries}

% ======== Motivate DSN types ==========
% For traffic reconstruction and forecasting in this study, we selected an informative set of external factors that have been shown to influence traffic dynamics~\cite{Ramhormozi22}. These factors include environmental data (weather and air quality), static road semantics (e.g., number of lanes, maximum speed), and positional data (coordinates and regional descriptions). Together, these factors form a comprehensive traffic observation.

For traffic reconstruction and forecasting, we incorporated additional environmental, semantic, and positional factors as contextual data. Together, these factors create a comprehensive view of traffic dynamics.

% For this study, we selected a set of external factors that have shown in the past to have a large influence on traffic dynamics \cite{Ramhormozi22}.
% The focus is on environmental data (weather and air quality), road semantics that describes the street leg with static values (e.g. lane number, max speed, etc.), and positional data in the form of coordinates and a regional description.
% All of them formed a traffic observation.

\begin{figure*}
    \centering
    \includegraphics[width=\textwidth]{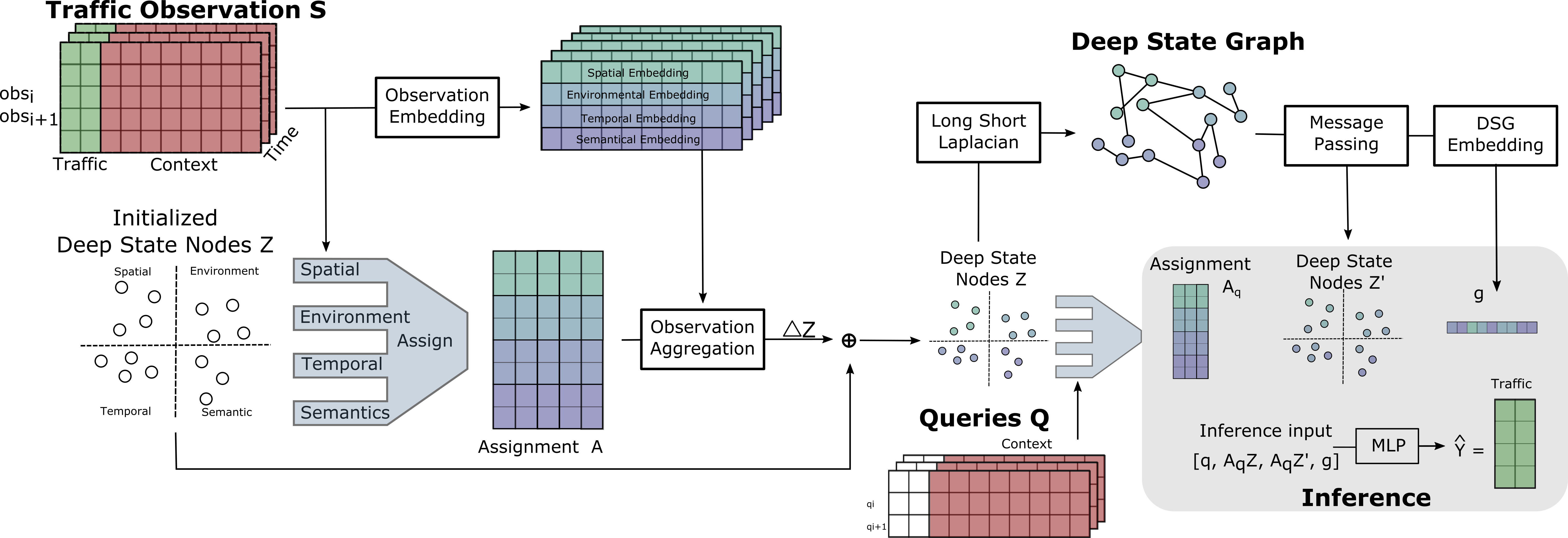}
    \caption{Overall DeepStateGNN Architecture.}
    \label{fig:dsgnn-arch}
\end{figure*}
% ======== What is a traffic observation? ==========
{\bf Traffic Observation:} \label{traffic-obs-def} A traffic observation consists of a combination of traffic measurements (including speed and flow) and contextual information at a specific sensor location during a defined time window. 
% The traffic measurements include average speed and flow, where flow is the number of vehicles passing a sensor in a 5-minute time period.
% while the contextual information is divided into static and time-dependent factors. 
% Static factors encompass spatial properties of the sensor and road semantics. Time-dependent factors include weather and air quality measurements at the sensor location during the time window.

{\bf Deep State Node (DSN):} Traffic sensors that exhibit similar characteristics can be grouped and represented through Deep State Nodes (DSNs). 
% The similarity among traffic sensors is defined by various metrics, resulting in different types of DSNs. 
In this work, we consider four DSN types: Spatial DSNs group sensors based on their geographical properties (e.g., neighborhood). Semantic DSNs cluster sensors according to road features (e.g., maximum lane speed). Environmental DSNs group sensors by similarity of weather and air quality conditions at their locations. % during the input window. 
Temporal DSNs cluster sensors based on observed traffic patterns. 
% Each DSN type can include multiple nodes, representing different characteristics (e.g., 3 nodes for sensors on roads with maximum lane speeds of 40, 65, and 75 MPH).

% ======== DSN grpah ==========
{\bf Deep State Graph (DSG):} A Deep State Graph is a compact, fixed-size graph representing the traffic network. 
Nodes in the DSG correspond to DSNs, each maintaining a latent state based on the aggregated observations of the sensors they represent. Edges in the DSG capture the short- and long-term similarities between these latent states. %, providing a compressed representation of the overall traffic conditions and interactions across the road network.

\textbf{Traffic Forecasting and Reconstruction Problems:} Given a window $k$ of traffic observations of length $W$, denoted as $X^{(k)}=(x_1^{(k)}, \ldots, x_W^{(k)}) \in \mathbb{R}^{|S^{(k)}| \times W \times F}$, where each $x_i^k$ represents a traffic observation with $F$ features (traffic and contextual) for a subset of sensors $S^{(k)}$ in the road network, as well as a window of query observations $X_q^{(k)}=(x_{q_1}^{(k)}, \ldots, x_{q_W}^{(k)}) \in \mathbb{R}^{|Q^{(k)}| \times W \times (F-2)}$ providing only the contextual values (missing speed and traffic flow measurements) for a query set of sensors $Q^{(k)}$ that don't overlap with $S$,
% $Q^{(k)}$ with $Q^{(k)} \cap S^{(k)} = \emptyset$ (missing speed and flow traffic measurements),
the task is to:\\
% (I) \textbf{Traffic Forecasting:} Predict future traffic measurements (speed and occupancy) over a horizon $H$, with $Q(k) \subset S(k)$ and $X^{(k)}$ being the input, denoted as $\hat{Y}_{Forecast}^{(k)}=(\hat{y}_{W+1}^{(k)}, \ldots, \hat{y}_{W+H}^{(k)}) \in \mathbb{R}^{|Q^{(k)}| \times H \times 2}$.
% , for a separate subset of sensors called query sensors $Q^{(k)}$. 
(I) \textbf{Traffic Reconstruction:} Reconstruct the traffic measurements for the last $H \in \mathbb{N}^{\leq W}$ timestamps of the input window for the query set $Q^{(k)}$, represented as $\hat{Y}^{(K)}_{Reconstruct}=(\hat{y}_{W-H+1}, \ldots, \hat{y}_W) \in \mathbb{R}^{|Q^{(k)}| \times H \times 2}$.\\
(II) \textbf{Traffic Forecasting:} Predict future traffic measurements over a horizon $H$ 
% without having traffic measurements 
for the query sensors $Q(k)$ 
% in the observation time window $W$, 
represented as $\hat{Y}^{(K)}_{Forecast}=(\hat{y}_{W+1}, \ldots, \hat{y}_{W+H}^{(k)}) \in \mathbb{R}^{|Q^{(k)}| \times H \times 2}$. 

To align with the goal of generalizing the traffic state for queries involving unknown sensors, we intentionally avoid the simpler forecasting scenario, where the traffic values of the query sensors are part of the input. Instead, we present our forecasting approach (Definition II) as a more challenging and realistic problem setting.
For simplicity of expressions, we drop the $(k)$ superscript for the rest of the paper.

\section{Proposed Model (DeepStateGNN)}
DeepStateGNN addresses the scalability and flexibility challenges of previous GNN-based traffic forecasting and reconstruction methods. The architecture, shown in Figure~\ref{fig:dsgnn-arch}, constructs a Deep State Graph (DSG) composed of a fixed number of Deep State Nodes (DSNs) representing sensor groups and their relationships. Graph Convolution is applied to the DSG to capture inter-node relationships in the DSN embeddings. These embeddings are then used to infer traffic conditions for query locations.

% \section{Proposed Model (DeepStateGNN)}
% Our proposed DeepStateGNN addresses the flexibility and scalability issues of previous GNN-based traffic forecasting/reconstruction approaches. Figure~\ref{fig:dsgnn-arch} shows the overall architecture.
% We begin by constructing a Deep State Graph (DSG), which consists of a fixed number of Deep State Nodes (DSNs) to represent the overall traffic state. Each traffic observation is assigned to a subset of DSNs using a combination of hand-crafted and learned metrics. The assigned observations are then aggregated for each DSN to create a latent representation.
% The relationships between DSNs are dynamically updated through the edge weights in the DSG, followed by applying Graph Convolution to capture inter-node correlations. 
% % Importantly, graph convolution is applied only to the DSG nodes, not the individual DSN nodes, which helps account for the speed and efficiency of our approach. 
% For any query location, relevant DSN embeddings from the convolved DSG are retrieved and used to infer the traffic conditions.
% at that location.

\subsection{Deep State Graph Construction}
The DSG represents the latent traffic state through DSNs that aggregate traffic observations from multiple sensors in a latent space. DSN states are initialized using an MLP based on the global input context, including the time of day and day of the week in our tasks. A combination of hand-crafted and learned metrics is employed to compute non-exclusive assignments from observations to DSNs. The rationale for using different types of assignments is that for some DSNs, such as spatial nodes, the assignment of sensors is straightforward based on spatial coverage. In contrast, the assignment for other DSNs, like environmental ones, is not predetermined and requires learning from data. After the assignment, the observations are embedded and aggregated to update the DSN states. Finally, the weighted and directional edges of the DSG are inferred based on long- and short-term relationships between DSN states. These steps are detailed in the following sections.

\subsection{Observation Assignments}\label{met:assignment}
\subsubsection{Static Assignments to Spatial and Semantic DSNs} 
Observation assignments to spatial and semantic DSNs are static, as the spatial and semantic properties of each traffic sensor remain unchanged during the observed window. 
For spatial DSNs, one DSN is allocated per neighborhood and freeway. 
Assignments are based on the distance of the traffic sensor to the neighborhood center (soft assignment) or a binary assignment indicating whether the sensor is located on the specific road the DSN represents. 
For semantic DSNs, nodes are allocated for each semantic property, such as the number of lanes or maximum speed (e.g., one DSN for a max speed of 40 MPH). 
Observations are assigned to these nodes based on binary criteria reflecting the semantic properties of the road that the sensor that recorded the observation is placed on.

\subsubsection{Dynamic Assignments to Environmental and Temporal DSNs}
DSG includes environmental and temporal DSNs that group traffic observations from sensors based on environmental factors (in our case weather and air quality) and temporal patterns. 
Sensors recording observations under similar conditions (e.g., same precipitation level) or showing similar traffic patterns should be mapped to the same environmental/temporal DSNs. 
% Other than static DSN assignment such a grouping affords external knowledge over the scope of an ordinary application.
Since the clusters corresponding to these DSNs (e.g., ``rainy'' environment) and the assignments from sensors to DSNs cannot be predefined, they must be learned from data. 
To achieve this, we specify the number of such DSNs as a hyperparameter and initialize their states using the same MLP as other DSNs. 

% For the assignment, we are learning a gating-based mechanism $gate$ (implemented as an MLP) for each DSN type that assigns soft weights from each observation to each DSN with $Z_{type}$ being the set of all DSNs from a certain type.
% The input takes those features $X_{f}$ that are relevant to the DSN type (e.g., all weather features for weather DSNs).
% The assignment function we utilize is as follows:
The assignment process involves learning a gating-based mechanism, implemented as an MLP, that assigns soft weights from each observation to each DSN. For a given DSN type (denoted as $Z_{type}$), the input features $X_{f}$, relevant to that DSN type (e.g., all weather-related features for environmental DSNs), are used in the assignment. The assignment function is defined as follows:
\begin{equation}
    a_i = gate([X_{f}; z_i]), \forall z_i \in Z_{type}
\end{equation}

Here, $z_i$ represents the state vector of a single DSN, and $[\cdot ; \cdot]$ denotes the concatenation operation. Each observation in the set $X_{f}$ is assigned an assignment vector $a_i \in \mathbb{R}^{|Z_{type}|}$. Concatenating all the assignment vectors forms the assignment matrix $A_{type} \in \mathbb{R}^{|Z_{type}| \times |S|}$ for the specific DSN type, where $|S|$ is the number of observation sensors.

% With $z_i$ being a single DSN node state after initialization and $[.;.]$ denoting the concatenation operation. Each observation in the set $X_{f}$ is assigned with a gating value forming the output vector $a_i \in \mathbb{R}^{|S|}$. With the concatenation of all $a_i$ an assignment matrix for a specific type is formed $A_{type} \in \mathbb{R}^{|Z_{type}| \times |S|}$

After each DSN type generates its type-specific assignment matrix for both static and dynamic types, these partial matrices are concatenated into a single large matrix.
The output of the assignment step is a tensor $A \in \mathbb{R}^{N \times |S|}$, where $N$ is the number of DSNs.
To enhance robustness against noise, we apply a threshold to discard assignments with very small weights.

\subsection{DSN State Update}
\subsubsection{Observation Embedding}\label{subsec:embed}
As described in Section \ref{traffic-obs-def}, traffic observations $X$ include traffic measurements over a window of time ($X_{traffic}$), static context such as road semantics ($X_{static}$), and dynamic context such as weather measurements during the same window ($X_{dynamic}$). Embedding these into a fixed-length space captures intra-series correlations efficiently. This also eliminates the need for separate graph instances for each timestamp, thereby saving memory and computation time.
% To effectively and efficiently process this information, each observation needs to be embedded into a fixed-length embedding space. 
% This ensures the capture of intra-series correlations across sequences of traffic and dynamic context measurements.
% As a result, we do not need a separate graph instance for each timestamp in the input window, which saves memory and computation time.

To embed observations, we first generate the dynamic context \( H_{dynamic} \) by processing the time-dependent features \( X_{dynamic} \) through a GRU:
\begin{equation}\label{eq:dynamic_context}
    H_{dynamic} = \text{GRU}(X_{dynamic}) \in \mathbb{R}^{|S| \times d_d}
\end{equation}

We then concatenate \( H_{dynamic} \) with static features \( X_{static} \) and pass it through an MLP to obtain the context embedding \( H_{context} \):

\begin{equation}\label{eq:context_embed}
    H_{context} = \text{MLP}([H_{dynamic}; X_{static}]) \in \mathbb{R}^{|S| \times d_c}
\end{equation}

The traffic sequence \( X_{traffic} \) is similarly embedded using a GRU, and the result \( H_{traffic} \) is combined with \( H_{context} \) to produce the final observation embedding \( H_{obs} \):

\begin{equation}
    H_{traffic} = \text{GRU}(X_{traffic}) \in \mathbb{R}^{|S| \times d_t}
\end{equation}
\begin{equation}
    H_{obs} = \text{MLP}([H_{traffic}; H_{context}]) \in \mathbb{R}^{|S| \times d_e}
\end{equation}

Here, \( d_d \), \( d_c \), \( d_t \), and \( d_e \) are the output dimensions of the GRU and MLP layers.

% To achieve this, DeepStateGNN employs an embedding module with the following steps. First, the time-dependent and static contextual values are embedded into $H_{context}$ by creating a fixed-size latent representation $H_{dynamic}$ from the time-dependent features using a GRU.
% The concatenation of $H_{dynamic}$ with the static features is used to calculate the context embedding with an MLP.

%     \begin{equation}\label{eq:dynamic_context}
%         H_{dynamic} = \text{GRU}(X_{dynamic}) \in \mathbb{R}^{|S| \times d_d}
%     \end{equation}
%     \begin{equation}\label{eq:context_embed}
%         H_{context} = \text{MLP}([H_{dynamic}; X_{static}]) \in \mathbb{R}^{|S| \times d_c}
%     \end{equation}
%     where $d_d$ is the hidden dimension of the GRU, $[\cdot ; \cdot]$ denotes concatenation and $d_c$ is the dimensionality of the output of MLP.
% Finally, the traffic sequence is embedded $H_{traffic}$ similar to the dynamic context and is concatenated with the embedded context and passed through another MLP to produce the final embedding:
%     \begin{equation}
%         H_{traffic} = \text{GRU}(X_{traffic}) \in \mathbb{R}^{|S| \times d_t}
%     \end{equation}
%     \begin{equation}
%         H_{obs} = \text{MLP}([H_{traffic}; H_{context}]) \in \mathbb{R}^{|S| \times d_e}
%     \end{equation}
%     where $d_t$ is the dimensionality of the GRU output, and \( d_e \) is the dimensionality of the final embedding. $H_{obs}$ contains the observation embeddings which will be used later for updating the state of DSNs.

\subsubsection{Observation Aggregation and State Update}
In this step, we aggregate observation embeddings into the DSN representations they are relevant to. For each DSN $z_i \in Z$, let $A_i \in \mathbb{R}^{|S|}$ denote the $i$-th row of the thresholded assignment matrix $A$, which defines the assignment weights from sensors to this DSN. We first scale the observation embeddings $H_{obs}$ by these assignment weights:
\begin{equation}
H_i = A_iH_{obs} \in \mathbb{R}^{|S| \times d_e}
\end{equation}
This scaling step introduces non-linearity due to the thresholding on $A$. Next, we aggregate the scaled observation embeddings in $H_i$ to update the state of $z_i$. After evaluating various aggregation functions (e.g., mean, max, transformers, multi-head attention), we found that the mean function offered the best balance of performance and simplicity. Thus, the update for DSN $z_i$ is:
\begin{equation}
\Delta z_i = \frac{1}{|S_i|} \sum_{s_j \in S_i} H_{i,j}
\end{equation}
where $S_i = \{s_k \in S \mid A_{i,k} > 0\}$ is the set of sensors with non-zero assignment scores to $z_i$, and $H_{i,j}$ is their corresponding scaled observation embedding in $H_i$. Finally, the DSN states $Z$ are updated using a residual connection:
\begin{equation}
Z \leftarrow Z + \Delta Z
\end{equation}

\subsubsection{Long-Short Laplacian}
After enriching DSN states with traffic observations, it is essential to capture correlations between DSNs to leverage information from related sensors. For example, if a neighborhood experiences rainy weather, the DSN for that neighborhood should be strongly connected to the DSN representing sensors in similar rainy conditions. These correlations are represented as edges in the DSG and are modeled using two types of similarities:

\begin{itemize}
    \item \textbf{Dynamic Short-Term Dependencies $(L_s)$}: This matrix is derived from multi-head attention applied to the updated DSN states $Z$, capturing short-term dependencies among DSNs within the current time window:
    \begin{equation}
        L_s = \text{Multihead\_Attention}(\text{keys}=Z, \text{queries}=Z)
    \end{equation}
    \item \textbf{Static Long-Term Similarities $(L_l)$}: This matrix captures long-term relationships between DSNs by learning source and target embeddings $E_s$ and $E_t$ for each node across the entire dataset, where $E_s, E_t \in \mathbb{R}^{N \times e}$ and $e$ is the embedding dimension. The long-term similarities are then computed as the softmax of the dot product between node embeddings in $E_s$ and $E_t$:
    \begin{equation}
        L_l = \text{softmax}(\text{relu}(E_s \times E_t^T))
    \end{equation}
\end{itemize}
The final Laplacian matrix $L$, termed the Long-Short Laplacian, combines these two matrices as follows:
\begin{equation}
    L = \alpha L_s + (1-\alpha) L_l
    \end{equation}
where $\alpha$ is a learnable parameter. This allows the model to balance short-term and long-term dependencies based on the downstream task. Finally, to enhance sparsity and reduce noise, the smallest $K\%$ of elements in $L$ are pruned.

\subsection{DSG Embedding}
With the updated DSN states $Z$ and adjacency matrix $L$, we form the DSG $G = (Z, L)$ for the given observation window. This graph is then processed through a modified Graph Convolutional Network (GCN)~\cite{kipf2017semisupervised}, which omits normalization and includes residual connections to preserve directed relationships and mitigate oversmoothing~\cite{chen2020measuring}. The output is the node embeddings $Z' \in \mathbb{R}^{N \times k}$ ($k$ is the embedding dimension).

Post-convolution, DSN states capture latent traffic information at the sensors within their coverage. To summarize the traffic state across the road network, we apply hierarchical pooling: first, a mixed mean-max-pooling for each DSN type, followed by another pooling to generate a final vector representation $g \in \mathbb{R}^{2k}$ for the entire DSG.

The states $Z$, $Z'$, and $g$ represent the traffic state at each DSN, the enriched state considering related DSNs, and the overall state across the network, respectively. We use them for forecasting/reconstruction at query sensor locations.

\subsection{Traffic Inference}
In this step, we infer the traffic measurements for the given query set $Q$. 
A query $q\in Q$ includes all the non-traffic observation features described in Section \ref{traffic-obs-def}. 
To infer the traffic for this query, we first use the assignment functions described in Section \ref{met:assignment} to find an assignment $A_q \in \mathbb{R}^{1 \times N}$ from this query to the DSN states. 
Next, we concatenate the query $q$, the weighted DSNs prior to convolution ($A_qZ$) for DSN-specific context, the weighted embedding of DSNs after convolution which is enriched with related DSNs' information ($A_qZ'$), and the global DSG embedding ($g$) for a global view of the traffic network. 
The global embedding captures additional information that cannot be passed through the local embeddings.
This could be an event happening on a different type of road close to the query that might not be well represented solely through the local embedding. 
This multi-view embedding allows for a comprehensive representation to infer the traffic measurements.
We then pass this representation to an MLP to perform the forecasting or reconstruction in a single shot for all the horizon timestamps:
\begin{equation}
    \hat{Y}_{traffic} = \text{MLP}([q; A_qZ; A_qZ'; g]) \in \mathbb{R}^{|Q|\times H \times 2}
    \label{eq:reconstruction}
\end{equation}

% John read to here 2024-08-14

\section{Experimental Evaluation}
\label{sec:experiment}

\subsection{Experimental Setup}
\subsubsection{METR-LA+ Dataset}

Traditional traffic benchmark datasets, such as METR-LA and PEMS-BAY, are constrained by their focus on a limited number of freeway sensors with curated data, where all sensors have uninterrupted observations across all timestamps. To more accurately capture real-world traffic conditions, we introduce \textbf{METR-LA+} \footnote{Due to anonymity, the link is omitted in this version but will be provided in the final paper.}. METR-LA+ broadens sensor coverage beyond its predecessors by including both freeway and arterial road sensors, while also offering contextual sensor data. To preserve the dataset’s real-world characteristics, no preprocessing was applied to address missing data, thereby retaining the natural occurrence of sensor outages. Further details can be found in Appendix A.

\begin{table}[]
    \centering
    % \tiny
    \caption{Summary of the time span, sample count, and average number of sensors per sample for each dataset split for both dataset variants: ``freeways-only'' and ``all-roads.''}
    \label{tab:dataset}
    \begin{adjustbox}{width=\linewidth, center}
    \begin{tabular}{l|c|c|c|c|c}
            && \multicolumn{2}{c|}{freeways-only} & \multicolumn{2}{c}{all-roads} \\
         Dataset&Time Window&Samples&Sensors&Samples&Sensors\\
         \hline
         Training &17.11-29.11.22&628&4671&3714&210\\
         Validation&29.11-06.12.22& 157&4511&928&193\\
         Test &06.12-16.12.22&339&4008&1319&207\\
    \end{tabular}
    \end{adjustbox}
    % \caption{Summary of the time span and sample count for each split of our dataset, along with the average number of sensors per sample, presented for both dataset variants: one covering only freeways and the other including all roads.}
\end{table}

\subsubsection{Dataset Configuration}
We assess two scenarios using METR-LA+: the ``freeways-only'' scenario, which includes only freeway data, and the default ``all-roads'' scenario. Each sample in the dataset comprises a 12-timestamp input window, with each timestamp representing a 5-minute interval, covering a total of 1 hour of data. To ensure fair performance evaluation, we include only samples with at least 1,000 valid sensor recordings per timestamp. Details on the dataset splits are provided in Table \ref{tab:dataset}. 
%Due to the higher number of arterial sensors and their greater likelihood of failure, the "all-roads" scenario produces more samples but with fewer valid sensors per sample. 
For baseline models, missing sensor data is filled with zeros.
% We evaluate two scenarios using METR-LA+: the ``freeways-only'' scenario, which includes only the freeway data, and the default ``all-roads'' scenario. Each sample in the dataset consists of a 12-timestamp input window (each timestamp representing a 5-minute interval), covering 1 hour of data. 
% To ensure fair performance evaluation, we include only samples with at least 1,000 valid sensor recordings per timestamp. Table \ref{tab:dataset} provides details on the dataset splits. 
% Due to the higher number of arterial sensors and their increased likelihood of failure, the ``all-roads'' scenario yields more samples but with fewer valid sensors per sample. For baseline models missing sensors are zero-filled.
%that require a complete sensor graph, each sample is represented as a tensor, with missing sensor data filled in with zeros to meet input requirements.

\subsubsection{Evaluation Setup}
For each sample, 90\% of the sensors are selected as observation sensors, which include both traffic measurements and contextual data in the input window. The remaining 10\% are designated as query sensors, providing only contextual data in the input window. We evaluate all baseline models and our approach on traffic reconstruction and forecasting over a horizon of 12 timestamps (60 minutes)\footnote{Experiments with horizons of 6 timestamps (30 minutes) and 3 timestamps (15 minutes) showed consistent relative performance, so we report results for the 60-minute horizon only.}, as discussed in Section~\ref{sec:preliminaries}. Performance is measured using established error metrics: Mean Absolute Error (MAE), Root Mean Squared Error (RMSE), and Mean Absolute Percentage Error (MAPE). Details on hardware, software specifications, and hyperparameters can be found in Appendices B and C.

\subsubsection{Baselines}
We compare our approach against several state-of-the-art time-series forecasting baselines. For fixed sensor graphs, we use DCRNN \cite{li2018} and A3tGCN \cite{Bai21}. Additionally, we include BysGNN \cite{Hajisafi23}, which combines a fixed sensor graph with meta-nodes that describe the state of similar nodes. For non-sensor graph baselines, we utilize SUSTeR \cite{woelker23}. Details on these baselines can be found in Section~\ref{sec:related-work}.

\subsubsection{Training}

To train DeepStateGNN, we utilize two training objectives aimed at reducing overfitting and guiding the model to learn effective embeddings for DSNs. The training objectives are as follows:
\begin{itemize}
    \item \textbf{Query Inference Loss ($L_1$)}: This is our primary objective, which minimizes the mean squared error (MSE) of normalized and combined traffic measurements (speed and flow) for each query sensor.
    %in the given downstream task (traffic forecasting or reconstruction).
    \item \textbf{Observation Reconstruction Loss ($L_2$)}: This secondary objective regularizes model training by encouraging the model to learn robust Deep State Node embeddings. After constructing the Deep State Graph, we use the observations as queries and attempt to reconstruct their traffic features (which are already available in the input) with minimal error, by minimizing the reconstruction error through MSE.
\end{itemize}
We combine these objectives and train the model using the final training objective $L$ as follows:
\begin{equation}
    L = L_1 + 0.9^{\pi} * \gamma L_2
\end{equation}
where $\gamma$ is a hyperparameter and $\pi$ is the current epoch number. This approach assigns greater weight to $L_2$ at the beginning of training to enhance embedding learning.

\subsection{Performance Evaluation}

\begin{table*}[]
    \centering
    \caption{Performance comparison between DeepStateGNN and baseline models for traffic forecasting and reconstruction tasks across the ``freeways-only'' and ``all-roads'' dataset scenarios. Results are reported as the mean $\pm$ standard deviation over 3 runs. The lowest error in each setup is highlighted in \textbf{bold}, while the second-best error is \underline{\emph{underlined}}. The "Improvement" row indicates the percentage improvement of DeepStateGNN over the next best baseline for each metric and task.}

    \label{tab:performance}
    \begin{adjustbox}{width=\linewidth, center}
    \begin{tabular}
{c|c|rr|rr|rr|rr} 
\multicolumn{2}{c}{}&\multicolumn{4}{c}{\textbf{freeways-only}}&\multicolumn{4}{c}{\textbf{all-roads}}\\
\small{Baseline}&\small{Metric}&\multicolumn{2}{c}{\small{Reconstruction}}&\multicolumn{2}{c}{\small{Forecast}}&\multicolumn{2}{c}{\small{Reconstruction}}&\multicolumn{2}{c}{\small{Forecast}}\\
\toprule
 &  & \small{Speed} & \small{Flow} & \small{Speed} & \small{Flow} & \small{Speed} & \small{Flow} & \small{Speed} & \small{Flow} \\ 
\toprule 
 \multirow{3}{*}{\small{A3TGCN}}& RMSE&$11.77\pm 0.06$&$5.82\pm 0.21$&$12.02\pm 0.23$&$6.01\pm 0.33$&$8.48\pm 0.04$&\underline{\emph{5.21 $\pm$ 0.00}}&$9.16\pm 0.02$&\underline{\emph{5.91 $\pm$ 0.11}}\\ 
& MAE&$8.66\pm 0.15$&$3.82\pm 0.41$&$9.02\pm 0.28$&$4.19\pm 0.47$&$5.05\pm 0.14$&\underline{\emph{2.59 $\pm$ 0.01}}&$5.59\pm 0.10$&\underline{\emph{3.26 $\pm$ 0.24}}\\ 
& MAPE&$23.41\pm 0.96$&$142.32\pm 47.38$&$24.13\pm 2.01$&$166.38\pm 50.40$&$26.30\pm 1.11$&$113.06\pm 4.78$&$27.93\pm 0.18$&$142.75\pm 23.04$\\ 
\midrule 
 \multirow{3}{*}{\small{BysGNN}}& RMSE&$11.55\pm 0.27$&$6.20\pm 0.25$&$12.04\pm 0.64$&$6.03\pm 0.30$&$7.01\pm 0.02$&$6.41\pm 0.60$&$7.66\pm 0.22$&$6.71\pm 0.27$\\ 
& MAE&$8.31\pm 0.23$&$4.15\pm 0.24$&$8.65\pm 0.08$&$3.94\pm 0.11$&$4.58\pm 0.22$&$4.57\pm 0.71$&$4.69\pm 0.03$&$4.76\pm 0.38$\\ 
& RMSE&$22.47\pm 1.98$&$189.29\pm 6.78$&$24.58\pm 1.40$&$162.60\pm 36.51$&$25.63\pm 0.88$&$179.89\pm 44.82$&$25.60\pm 0.18$&$196.20\pm 18.29$\\ 
\midrule 
 \multirow{3}{*}{\small{DCRNN}}& RMSE&\underline{\emph{9.64 $\pm$ 0.68}}&\emph{\underline{4.88 $\pm$ 0.58}}&\underline{\emph{10.95 $\pm$ 0.70}}&\underline{\emph{5.47 $\pm$ 0.57}}&\underline{\emph{5.70 $\pm$ 0.09}}&$5.59\pm 0.51$&\underline{\emph{6.10 $\pm$ 0.28}}&$6.25\pm 0.39$\\ 
& MAE&\underline{\emph{6.74 $\pm$ 0.66}}&\underline{\emph{2.90 $\pm$ 0.66}}&\underline{\emph{7.82 $\pm$ 0.64}}&\underline{\emph{3.58 $\pm$ 0.61}}&\underline{\emph{3.94 $\pm$ 0.44}}&$3.03\pm 0.97$&\underline{\emph{4.23 $\pm$ 0.56}}&$3.83\pm 0.93$\\ 
& MAPE&\underline{\emph{17.56 $\pm$ 1.67}}&\underline{\emph{107.22 $\pm$ 44.74}}&\underline{\emph{20.89 $\pm$ 2.50}}&\underline{\emph{142.74 $\pm$ 41.17}}&\underline{\emph{23.47 $\pm$ 0.01}}&$140.81\pm 60.56$&\underline{\emph{24.55 $\pm$ 0.46}}&$165.50\pm 68.44$\\ 
\midrule
 \multirow{3}{*}{\small{SUSTeR}}& RMSE&$12.49\pm 0.05$&$6.83\pm 0.05$&$12.65\pm 0.10$&$6.80\pm 0.17$&$9.01\pm 0.04$&$5.83\pm 0.05$&$9.62\pm 0.02$&$6.17\pm 0.08$\\ 
& MAE&$8.81\pm 0.00$&$4.61\pm 0.01$&$8.91\pm 0.01$&$4.61\pm 0.10$&$4.88\pm 0.02$&$3.31\pm 0.03$&$5.18\pm 0.01$&$3.52\pm 0.07$\\ 
& MAPE&$26.35\pm 0.43$&$230.72\pm 2.09$&$26.68\pm 0.63$&$221.00\pm 8.01$&$26.42\pm 0.17$&${\bf 92.30\pm 0.45}$&$27.05\pm 0.04$&${\bf 105.85\pm 1.04}$\\ 
\midrule 
 \multirow{3}{*}{\small{DeepStateGNN}}& RMSE&${\bf 8.47\pm 0.06}$&${\bf 4.06\pm 0.04}$&${\bf 9.20\pm 0.19}$&${\bf 4.35\pm 0.10}$&${\bf5.21\pm 0.00}$&${\bf 4.89\pm 0.01}$&${\bf 5.55\pm 0.01}$&${\bf 5.52\pm 0.02}$\\ 
& MAE&${\bf 5.94\pm 0.08}$&${\bf 2.34\pm 0.04}$&${\bf6.37\pm 0.14}$&${\bf2.56\pm 0.12}$&${\bf3.48\pm 0.01}$&${\bf 2.38\pm 0.03}$&${\bf 3.70\pm 0.02}$&${\bf 2.96\pm 0.05}$\\ 
& MAPE&${\bf 15.12\pm 0.26}$&${\bf 79.03\pm 5.30}$&${\bf 17.00\pm 0.16}$&${\bf 84.85\pm 12.26}$&${\bf 21.36\pm 0.04}$&\emph{\underline{104.44 $\pm$ 2.40}}&${\bf 22.09\pm 0.08}$&\underline{\emph{112.54 $\pm$ 4.69}}\\ 
\midrule\midrule
% \bottomrule

 \multirow{3}{*}{\small{Improvement(\%)}}& RMSE&$12.14$\%&$16.80$\%&$15.98$\%&$20.48$\%&$8.60$\%&$6.14$\%&$9.02$\%&$6.60$\%\\ 
& MAE&$11.87$\%&$19.31$\%&$18.54$\%&$28.49$\%&$11.68$\%&$8.11$\%&$12.53$\%&$9.20$\%\\ 
& MAPE&$13.90$\%&$26.29$\%&$18.62$\%&$40.56$\%&$8.99$\%&*&$10.02$\%&*\\ 

    \end{tabular}
    \end{adjustbox}
    \end{table*}

Table~\ref{tab:performance} presents the performance results of DeepStateGNN and baseline models for traffic reconstruction and forecasting tasks on the METR-LA+ dataset. Our proposed DeepStateGNN consistently outperforms the baselines across all settings, with the exception of MAPE for traffic flow forecasting and reconstruction in the ``all-roads'' scenario. Notably, DeepStateGNN achieves up to 40\% improvement in traffic flow and 16\% in traffic speed for forecasting and reconstruction metrics, respectively. These results highlight the effectiveness of our DeepState Graph representation, which groups similar sensors based on contextual factors to accurately reconstruct or forecast traffic at unknown sensor locations from sparse observations.

A key observation is that the performance improvement is more significant in traffic forecasting than in traffic reconstruction for the same metrics. This difference can be attributed to the difficulty of the forecasting task, which requires predicting future values without knowing any of the prior traffic observations, whereas reconstruction estimates values for the current time window for which some observations are available. This suggests that DeepStateGNN is particularly well-suited for more complex scenarios, leveraging its high-level representation of traffic states for similar sensors.

Among the baselines, DCRNN performs best in the "freeways-only" scenario, consistent with its success in well-known benchmarks like METR-LA and PEMS-BAY. DCRNN’s use of a sensor graph based on road-network distance effectively captures spatial relationships critical to understanding traffic on highways, where signal propagation is strong~\cite{Pan22}. However, in the ``all-roads'' scenario, while DCRNN remains the best among the baselines (but outperformed by DeepStateGNN) for traffic speed forecasting and reconstruction, A3TGCN (for MAE and RMSE) and SUSTeR (for MAPE) outperform DCRNN in traffic flow forecasting and reconstruction. A3TGCN’s attention-based mechanism and SUSTeR’s abstract nodes allow these models to better capture the diverse patterns seen on arterial roads, where traffic does not necessarily follow the spatial propagation typical of highways.

To further explore the limitations of sensor-graph-based approaches, we compared DeepStateGNN with DCRNN as the ratio of query sensors increased. Initially, 10\% of sensors at each timestamp were used as queries, with the remaining sensors serving as observations. In the reconstruction task, as the query ratio increased to 80\% and 90\%, DCRNN’s MAE for the speed feature increases by 7\% and 12\%, respectively, and by 21\% and 87\% for traffic flow. In contrast, DeepStateGNN limited the error increase to less than 3\% for speed and 18\% for traffic flow. This demonstrates the robustness of DeepStateGNN, which, unlike DCRNN, is not constrained by a sensor-graph, enabling it to effectively handle varying amounts of missing sensors.

\subsection{Computation Time}
To assess the scalability of DeepStateGNN, we compare its training time per epoch with that of baseline models using our ``all-roads'' dataset. For this analysis, we limit the number of sensors to subsets of 200, 1000, 3000, and 4000. Each experiment is repeated three times on an identical GPU with a batch size of 32.

\begin{figure}
    \centering
    \includegraphics[width=0.8\columnwidth]{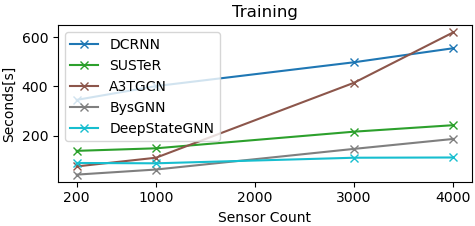}
    \caption{Training time per epoch with a constant batch size.}
    \label{fig:scaling}
\end{figure}

Figure \ref{fig:scaling} presents the training time results. DeepStateGNN shows the fastest training times for larger sensor sets.
For smaller datasets, the fixed size of the DeepStateGNN graph is similar to the sensor graph in other models, causing DeepStateGNN to perform less efficiently.
However, as the sensor count increases, DeepStateGNN demonstrates significant scalability advantages, maintaining a linear scaling pattern with a much slower growth rate than all other baselines, approaching near-constant behavior. This highlights DeepStateGNN's efficiency in handling larger datasets compared to other models.

% To evaluate the scalability of DeepStateGNN, we compare its training time per epoch with that of baseline models. We perform this comparison using our ``all-roads'' dataset to include all roads. To perform the analysis, we restrict the number of sensors to subsets of 200, 1000, 2000, and 4000. All models are trained using a batch size of 32, with each experiment running on a single GPU, repeated three times for consistency.

% Figure \ref{fig:scaling} illustrates the training time results. DeepStateGNN achieves the fastest training times for larger sensor sets, but for smaller sensor sets, A3T-GCN and BysGNN outperform DeepStateGNN. This discrepancy arises because, for smaller datasets, the size of the fixed DeepStateGNN graph is comparable to that of the sensor graph used in other models. Consequently, the overhead associated with assigning observations to DeepState nodes (DSNs) makes DeepStateGNN slower in these cases.

% However, DeepStateGNN demonstrates significant scalability advantages as the sensor count increases and maintains linear scaling with a much slower growth rate than all other baselines, approaching near-constant behavior. This demonstrates DeepStateGNN's efficiency in handling larger datasets compared to the baseline models.

% \begin{figure}
%     \centering
%     \includegraphics[width=0.6\columnwidth]{AnonymousSubmission/LaTeX/figs/train_time.png}
%     \caption{Training time for one epoch with constant batch size}
%     \label{fig:scaling}
% \end{figure}

\subsection{Ablation Study}
To evaluate the contributions of different components in DeepStateGNN, we compare the full model ($DSG$) against four variants: removing dynamic DSNs ($DSG_{-d}$), static DSNs ($DSG_{-s}$), message passing ($DSG_{-gcn}$), and observation reconstruction loss ($DSG_{-L2}$). Table \ref{tab:ablation} reports error metrics for average speed reconstruction on the "all-roads" scenario. The full model outperforms the ablated variants in 9 out of 12 metric comparisons, with only minor differences observed in favor of $DSG_{-d}$ for MAPE (0.03\% lower), $DSG_{-gcn}$ for RMSE (0.01 lower), and a tie in MAE with $DSG_{-d}$. This underscores the importance of incorporating both static and dynamic DSNs and leveraging message passing for sensor group correlations.

While $DSG_{-d}$ shows only a slight performance decline, the configuration using only dynamic DSNs ($DSG_{-s}$) still significantly outperforms the best baseline (DCRNN) on the same task, indicating dynamic context like weather is important but that spatial features like neighborhood have a greater impact. The $DSG_{-L2}$ variant sees a notable 70\% drop in MAE, emphasizing the critical role of observation reconstruction loss in regularizing model training and improving generalization.

% \subsection{Ablation study}
% To thoroughly evaluate the contributions of the DSN nodes and the DeepState Graph (DSG) within the DeepStateGNN framework, we evaluate four variants by removing dynamic DSNs, removing static DSNs, removing the message passing, and removing the $L_2$ loss, having the names $DSG_{-d}$, $DSG_{-s}$, $DSG_{-id}$, and $DSG_{-L2}$ respectively. Table \ref{tab:ablation} shows the error metrics for the average speed (occupancy looking similar) for the reconstruction task on the "all roads" scenario. Our approach outperforms all variants except for $DSG_s$ on the MAPE metric showing the importance of including static and dynamic DSN types into a DSG which is subject to a message passing layer and the importance of our pretraining loss. Although $DSG_s$ performs equal in the MAPE with us a combination of both is relevant in optimizing MAE and RMSE.  

    \begin{table}[]
    \small
    \centering
    % \caption{Ablation study with a reconstruction on `all roads`. Reporting mean metrics about the average speed. Standard deviation is omitted due to negligible values.}
    \caption{Ablation study: Speed reconstruction on the ``all-roads'' scenario. Metrics represent the mean values of 3 runs, with standard deviation omitted due to negligible variation.}
    \begin{adjustbox}{width=\columnwidth, center}
    \begin{tabular}
{c|r|r|r|r|r} 
Metric & DSG & $DSG_{-d}$& $DSG_{-s}$& $DSG_{-gcn}$& $DSG_{-L2}$\\ 
\hline
MAE &5.21 &5.21&5.37&5.24&8.9\\ 
RMSE &3.48&3.51&3.72&3.47&4.9\\ 
MAPE &21.36\%&21.33\%&22.28\%&21.45\%&25.9\%\\ 

    \end{tabular}
    \end{adjustbox}
    \label{tab:ablation}
    \end{table}

% \section{Discussion}

\section{Conclusion}
In this work, we presented DeepStateGNN, a novel GNN framework that clusters traffic sensors into high-level nodes based on contextual similarity and observed patterns, forming a fixed-size DeepState graph. This approach addresses key limitations in previous GNN-based traffic forecasting methods, particularly in terms of scalability, flexibility, and effectiveness in handling incomplete traffic observations. Our extensive experiments on the new METR-LA+ dataset demonstrated that DeepStateGNN significantly outperforms state-of-the-art baselines in both traffic forecasting and reconstruction tasks, while also offering improved computational efficiency.

\bibliography{aaai25}

\end{document}